\documentclass[10pt,twocolumn,letterpaper]{article}

\usepackage{cvpr}  
\definecolor{cvprblue}{rgb}{0.21,0.49,0.74}
\usepackage[pagebackref,breaklinks,colorlinks,allcolors=cvprblue]{hyperref}

\usepackage{booktabs}
\usepackage{multirow}
\usepackage{algorithm}
\usepackage{algpseudocode}
\usepackage{amsmath}
\usepackage{amssymb}
\usepackage[table]{xcolor}
\usepackage{siunitx}

\usepackage{graphicx}

\title{Z-Order Transformer for Feed-Forward Gaussian Splatting}

\begin{document}
% \maketitle

\author{
Can Wang\textsuperscript{1} \quad
Lei Liu\textsuperscript{1} \quad
Wei Jiang\textsuperscript{2} \quad
Dong Xu\textsuperscript{1}\textsuperscript{$\dagger$} \\
\textsuperscript{1}The University of Hong Kong \quad \textsuperscript{2}Futurewei Technologies Inc.
}

\twocolumn[{%
\renewcommand\twocolumn[1][]{#1}%
\maketitle
\centering
\includegraphics[width=0.97\linewidth]{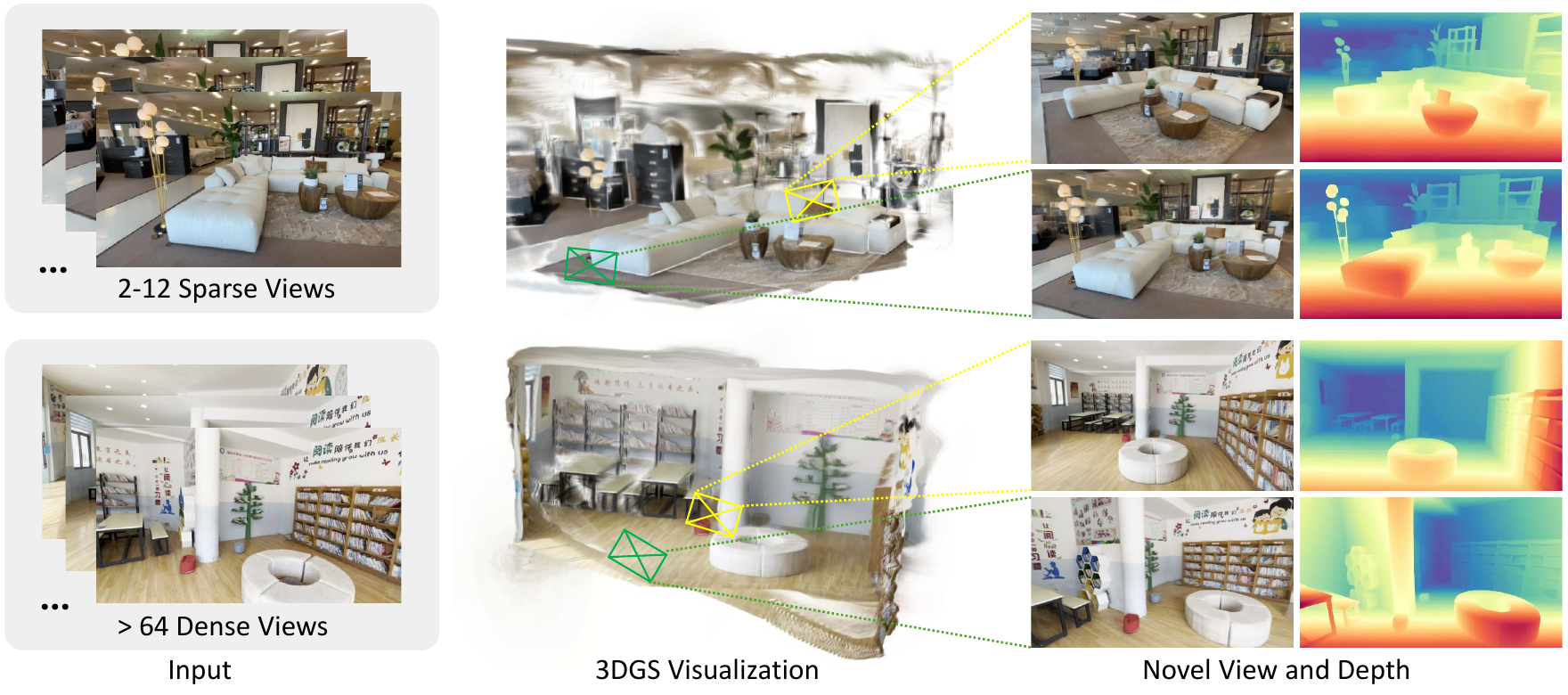}
\vspace{-1em}
\captionof{figure}{We propose a Z-order transformer that reconstructs 3D Gaussians from arbitrary multi-view captures for novel view synthesis. Compared with previous methods, our approach achieves more effective and efficient 3D Gaussian Splatting (3DGS) reconstruction while requiring fewer Gaussian primitives. 
\vspace{1em}
}
\label{fig:teaser}
}]
\footnote{\textsuperscript{$\dagger$}\ Dong Xu is the corresponding author.}

\begin{abstract}

Recent advances in 3D Gaussian Splatting (3DGS) have enabled significant progress in photorealistic novel view synthesis. However, traditional 3DGS relies on a slow, iterative optimization process, which limits its use in scenarios demanding real-time results. To overcome this bottleneck, recent feed-forward methods aim to predict Gaussian attributes directly from images, but they often struggle with the redundancy of Gaussian primitives and rendering quality. In this work, we introduce a transformer-based architecture specifically designed for feed-forward Gaussian Splatting.
Our key insight is that spatial and semantic relationships among Gaussians can be effectively captured through a sparse attention mechanism, enabled by a Z-order strategy that organizes the unstructured Gaussian set into a spatially coherent sequence. 
Furthermore, we incorporate this Z-order strategy to adaptively suppress redundancy while preserving critical structural details. 
This allows the transformer to efficiently model context, compress Gaussian primitives, and predict Gaussian attributes in a single forward pass. 
Comprehensive experiments demonstrate that our method achieves fast and high-quality novel view synthesis with fewer Gaussian primitives. 
\end{abstract}    
\section{Introduction}
\label{sec:intro}

Recent advances in 3D Gaussian Splatting (3DGS)~\cite{zhong2025generative,kerbl20233d,yu2024mip,ye2025gsplat,liu20264dgs} have demonstrated remarkable progress in fast reconstruction, real-time rendering, and high-quality novel view synthesis by representing a scene as a collection of anisotropic Gaussian primitives optimized from multi-view images. However, the original 3DGS suffers from limited generalization capacity and remains time-consuming, as it relies on per-scene gradient-based optimization to fit the Gaussian parameters.

To address these limitations, recent studies have introduced feed-forward GS~\cite{chen2024mvsplat,pixelsplat,xu2025depthsplat,liu2025monosplat,jiang2025anysplat,zhang2025flare,ye2025nopose,zheng2024gps}, which replaces the costly per-scene optimization with a learned prediction framework. Instead of iteratively adjusting Gaussian parameters through gradient descent, feed-forward GS employs a neural network to directly infer Gaussian attributes such as position, scale, opacity, and color from input images in a single forward pass.

\begin{figure}[htbp]
  \centering
  \includegraphics[width=0.41\textwidth]{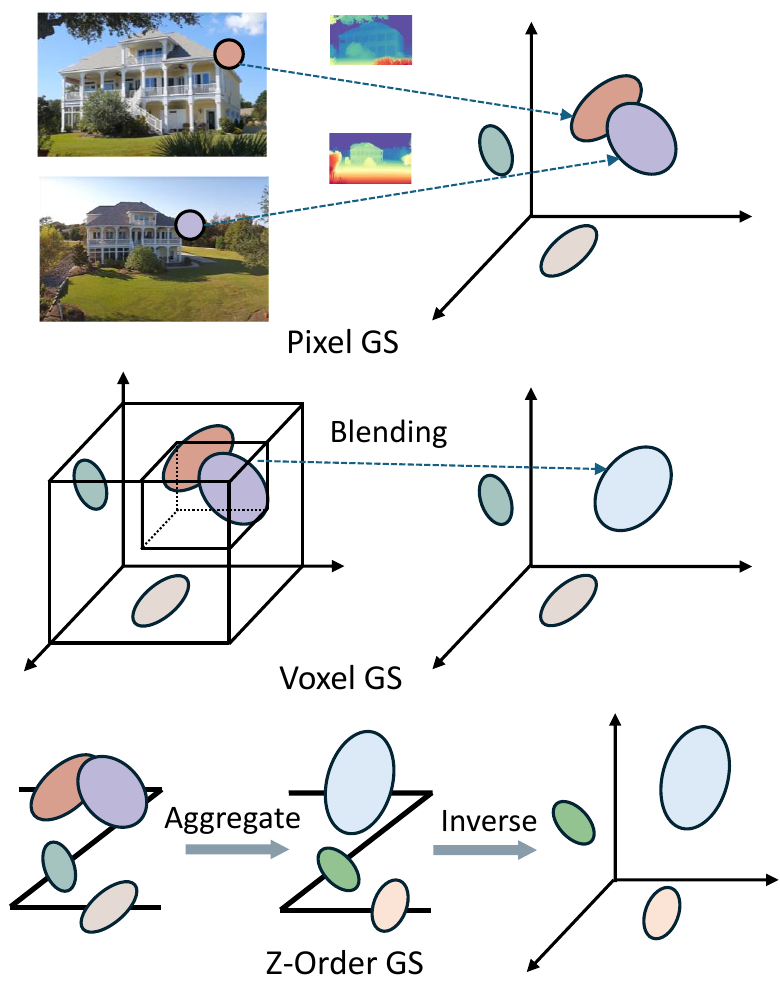}
  \vspace{-1em}
  \caption{Gaussian representations in feed-forward GS.}
  \label{fig:mot}
  \vspace{-1em}
\end{figure}

A key component of current feed-forward GS is the pixel-level Gaussian representation~\cite{chen2024mvsplat,pixelsplat,xu2025depthsplat,liu2025monosplat,zhang2025flare,ye2025nopose,zheng2024gps}, as illustrated in Fig.~\ref{fig:mot} (top). 
Each pixel in each input view is associated with one or multiple Gaussian primitives. The positions of these Gaussians are initialized by projecting the 2D pixel coordinates into 3D space through depth warping.
This design allows Gaussian parameters to be directly predicted from dense image features through a lightweight Gaussian head, typically implemented with a CNN. 
However, even associating one Gaussian per pixel can lead to a large number of primitives, which increases memory consumption and rendering costs, especially for high-resolution or multi-view inputs. For example, two 512×512 input views produce over 50W Gaussian primitives.

As illustrated in Fig.~\ref{fig:mot} (middle), voxel-based GS \cite{jiang2025anysplat,wanglearning,ren2024scube} mitigates this problem by aggregating neighboring pixel-level Gaussians within each voxel cell via a blending strategy.
While this method reduces redundancy, the discretization of 3D space inevitably introduces quantization errors that can blur fine details and sharp boundaries. Increasing the voxel resolution alleviates this problem but causes an exponential increase in memory usage and computational cost, as the number of voxels scales cubically with resolution. Furthermore, the fixed voxel grid is inefficient for representing irregular or sparse regions, leading to redundant empty cells and wasted computation.

We propose a Z-order \cite{morton1966computer} transformer to address the aforementioned issues.
As illustrated in Fig.~\ref{fig:mot} (bottom), our method first encodes pixel-level GS into a Z-order GS representation, which forms a compact 2D sequence while preserving spatial locality. This design removes the need for a dense voxel grid, resulting in higher computational and memory efficiency. The Z-curve ordering further enables fast neighborhood access and consistent spatial grouping, facilitating efficient GS aggregation and scalable feed-forward inference. Moreover, we propose a sparse attention mechanism consisting of group attention and top-k attention to fully exploit the locality of the Z-order GS representation while avoiding the high computational cost of full attention. 
After aggregation, the Z-order GS representation can be mapped back to 3D space through the inverse Z-order transformation. 
Considering the availability of dense views during inference and the resulting efficiency concerns, we further propose a Z-order-based Maximum Coverage Viewpoint Selection algorithm to effectively reduce view redundancy and improve efficiency.
Experimental results demonstrate that our Z-order transformer significantly improves efficiency and rendering quality while requiring fewer Gaussian primitives.

\section{Related Work}
\label{sec:related}

\begin{figure*}[htbp]
  \centering
  \includegraphics[width=0.98\textwidth]{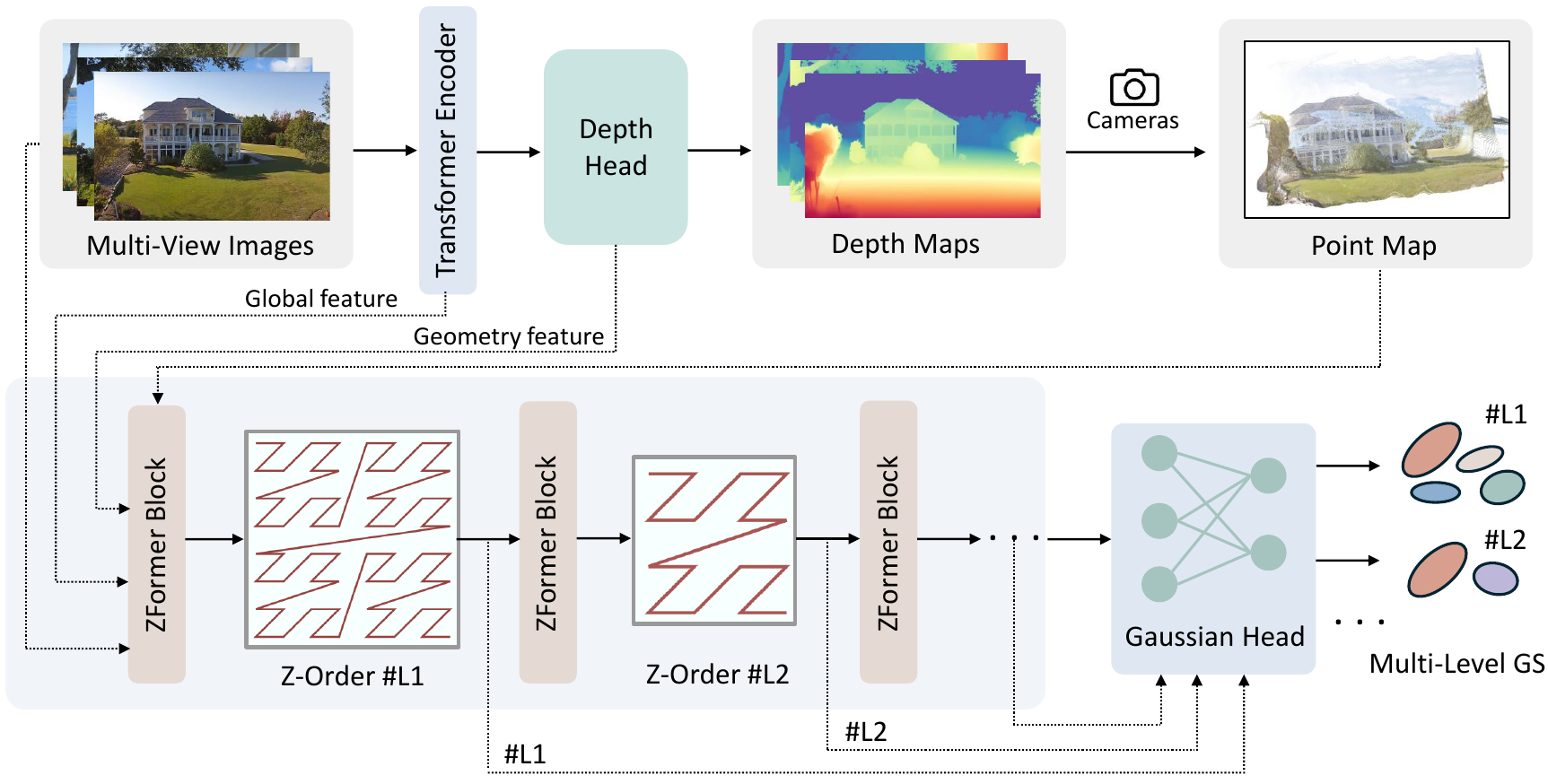}
  \vspace{-0.5em}
  \caption{\textbf{Framework.} Given multi-view images, our method first utilizes a transformer encoder and a depth head to generate depth maps, which are then projected into a 3D point map based on the camera. Next, global features are extracted from the transformer encoder, and geometry features are derived from the depth head. These features, along with the pixel color and point map, are processed through our ZFormer blocks to generate Z-order-based Gaussian representations. Finally, the Gaussian representations are passed to the Gaussian head, which generates multi-level GS for further rendering.}
  \label{fig:framework}
  \vspace{-1em}
\end{figure*}

\noindent\textbf{Per-scene Optimized Gaussian Splatting.} 
3D Gaussian Splatting (3DGS)~\cite{kerbl20233d} is an efficient representation for novel view synthesis that models a 3D scene as a collection of anisotropic Gaussian primitives. By directly optimizing these primitives through differentiable rendering, 3DGS achieves high-quality, photorealistic results. It has motivated many subsequent works~\cite{ye2025gsplat,cheng2024gaussianpro,zhang2024gaussian,kong2025rogsplat,li20253d,feng2025flashgs,yu2024mip} to improve its efficiency and rendering quality. For example, MipSplatting~\cite{yu2024mip} extends 3D Gaussian Splatting by introducing a multiscale, anti-aliased rendering formulation to address level-of-detail and aliasing issues, thereby enhancing rendering robustness and visual fidelity.
However, these optimization-based 3DGS methods are tailored to individual 3D scenes, lacking real-time reconstruction capability and generalization ability. To address this limitation, we propose a feed-forward 3DGS framework.

\noindent\textbf{Feed-forward Gaussian Splatting.} Feed-forward 3DGS employs a learned prediction network to directly infer Gaussian attributes from input images in a single forward pass. 
At present, most feed-forward GS methods~\cite{chen2024mvsplat,xu2025depthsplat,liu2025monosplat,zhang2025flare,ye2025nopose,zheng2024gps,pixelsplat} predict pixel-wise Gaussians based on pixel-aware features. For instance, PixelSplat~\cite{pixelsplat} and DepthSplat~\cite{xu2025depthsplat} predict 3D Gaussian primitives that are spatially aligned with pixel-level features from input images, achieving efficient and scalable novel view synthesis.
However, pixel-level-based methods often generate an excessive number of primitives, leading to increased memory consumption and rendering costs.
To overcome this limitation, voxel-based methods\cite{jiang2025anysplat,wanglearning,ren2024scube} have been proposed. For instance, AnySplat~\cite{jiang2025anysplat} adopts voxel-level techniques to optimize the processing of 3D Gaussian primitives through a differentiable voxelization module. However, the fixed voxel grid is inefficient for representing irregular or sparse regions, leading to numerous empty cells and redundant computations.
To address this, our method introduces a Z-order transformer that enables fast access to neighboring Gaussians and consistent spatial grouping, resulting in an efficient and high-quality feed-forward GS representation.

\noindent\textbf{Z-order Strategy.}
Z-order~\cite{morton1966computer} is a space-filling curve technique that converts multi-dimensional data into a low-dimensional sequence, with the goal of preserving spatial locality so that neighboring points in high-dimensional space remain adjacent in the lower-dimensional ordering.
Benefiting from its ability to preserve spatial locality, Z-order has been extensively utilized in spatial databases to enable efficient indexing and range queries~\cite{orenstein1989redundancy,orenstein1986spatial,lee2010z}, in computer graphics for texture mapping and hierarchical level-of-detail control~\cite{tarini2006ambient,lindstrom2001visualization}, and in point cloud processing for serialization and fast neighborhood retrieval~\cite{wu2024point,liang2024pointmamba,connor2010fast,wang2023octformer}.
For instance, Point Transformer V3~\cite{wu2024point} presents a simplified and scalable 3D backbone utilizing Z-order–based point serialization to enable efficient neighborhood aggregation, broader receptive fields, and superior speed and memory efficiency.
This motivates us to integrate the Z-order principle into 3DGS to enhance spatial coherence and achieve a more efficient and robust scene representation.

\section{Method}
\label{sec:method}

As shown in Fig.~\ref{fig:framework}, we propose a Z-order transformer that performs feed-forward 3DGS reconstruction for novel view synthesis given a set of multi-view image captures.
The structure of the following part is organized as follows: First, we provide an overview of the key concepts, including feed-forward 3D Gaussian prediction and the Z-order curve (\textbf{Sec. 3.1}). Next, we introduce the Z-order splat transformer (\textbf{Sec. 3.2}). We then present a multi-stage training strategy to train our framework (\textbf{Sec. 3.3}). Finally, we propose a Z-order-based maximum coverage viewpoint selection algorithm to reduce view redundancy and inference costs when dense captures are available (\textbf{Sec. 3.4}).

\subsection{Preliminaries}

\noindent\textbf{Feed-forward 3D Gaussian Splatting.}
Given $N$ views of a 3D scene $\mathbf{I}  =\left \{  I \right \}_{i=1}^N \in \mathbb{R}^{N\times 3\times H\times W}$, feed-forward 3DGS refers to the direct prediction of Gaussian primitives via a neural network $\mathcal{F}$, as follows:
\begin{equation}
\left\{ G_k:\left( \mu_k, \sigma_k, r_k, s_k, c_k \right) \right\}_{k=1}^{K}=\mathcal{F}(\left \{  I \right \}_{i=1}^N)
\end{equation}
Each Gaussian $G_k$ is characterized by a set of parameters:
$\mu_k\in \mathbb{R}^3 $ represents the mean of the $k^{th}$ Gaussian, $\sigma_k\in \mathbb{R}^+$ is the opacity, $r_k\in \mathbb{R}^4$ is the orientation quaternion, $s_k\in \mathbb{R}^3$ is the anisotropic scale, $c_k\in \mathbb{R}^{27}$ is the color represented by 2nd-order spherical harmonics (SH) coefficients. 
The key challenge of feed-forward 3DGS is to design a function $\mathcal{F}$ that is highly efficient and capable of predicting high-fidelity Gaussian primitives using as few Gaussian points as possible. Our work addresses this challenge with a novel Z-order transformer architecture that efficiently and effectively predicts compact yet high-quality Gaussian representations.

\noindent\textbf{Z-order Curve.}
The Z-order curve, also known as the Morton curve or Morton order, is a space-filling curve that maps multi-dimensional data onto a one-dimensional linear ordering while preserving spatial locality \cite{morton1966computer,wu2024point,zengzeta}. Let a 3D point be given as $P=(x,y,z)$. Each coordinate is expressed in binary using at most $depth=d$ bits:
\begin{equation}
    x = \sum_{i=0}^{d-1} x_i \, 2^i, \quad y = \sum_{i=0}^{d-1} y_i \, 2^i, \quad z = \sum_{i=0}^{d-1} z_i \, 2^i
\end{equation}
where $x_i, y_i, z_i \in \{0,1\}$ denotes the $i^{th}$ least significant bits.
The Z-order code is obtained by interleaving the bits of the three coordinates. Formally:
\begin{equation}
    \mathbf{Z}(x,y,z) = \sum_{i=0}^{d-1} \Big( x_i \cdot 2^{3i} \;+\; y_i \cdot 2^{3i+1} \;+\; z_i \cdot 2^{3i+2} \Big)
    \label{eq:zorder}
\end{equation}
This mapping is efficient for spatial applications as it maintains locality while providing a compact low-dimensional representation of multi-dimensional points \cite{wu2024point,zengzeta}.

\subsection{Z-Order Splat Transformer}

\noindent\textbf{Depth Estimation}. We begin by predicting the depth maps via a transformer encoder and a depth head, given the $N$ view input images $\mathbf{I}  =\left \{  I \right \}_{i=1}^N$. The structure of the network follows DINOv2-Small~\cite{oquabdinov2}, which patchifies $\mathbf{I}$ into tokens $\mathbf{t}
\in \mathbb{R}^{N\times l\times d}$, where $l=\frac{H\times W}{p}$ is the token number of one image, $p=14$ is the patch size, and $d=384$ is the token dimension. 
Then a DPT-based depth head~\cite{ranftl2021vision} is applied to the tokens $\mathbf{t}$ to predict the depth maps $\mathbf{D} \in \mathbb{R}^{N\times H\times W} $. Specifically, these tokens $\mathbf{t} $ will first be reshaped into feature maps and then multi-scale representations $\left \{ F_1,F_2,F_3,F_4 \right \}$ from different backbone layers are extracted, where $F_i\in \mathbb{R}^{N\times d_i\times H/p_i\times W/p_i}$ for $i \in \{1, 2, 3, 4\}$. Then these multi-scale feature maps will be fused using linear projections, upsampling, and transformer-style layers to progressively recover spatial resolution and obtain the geometry feature $F_{\text{geom}}\in \mathbb{R}^{N\times 64\times H\times W}$. Then convolutional layers are used to predict depth maps $\mathbf{D} $ from $F_{\text{geom}}$.
\vspace{0.5\baselineskip}

\noindent\textbf{Gaussian Point Representation}.
Formally, we denote the Gaussian point representation that will be used for the subsequent ZFormer block as $\mathbf{R}=\left \{\mathbf{P}, \mathbf{F},\mathbf{I}\right \} $, where $\mathbf{P}\in \mathbb{R}^{N\times (HW)\times 3}$ is obtained by projecting the depth $\mathbf{D} $ to point maps in world space via the camera of each input view. These points will be used to provide initialization for Gaussian centers.
$\mathbf{F}\in \mathbb{R}^{N\times 96\times H\times W}$ is the point-wise feature map, which will be used to predict Gaussian parameters. 
To obtain $\mathbf{F}$, we first reshape the tokens $\mathbf{t}$ into feature maps and then use two unsampling conv layers to obtain the global feature map $F_{\text{global}}\in \mathbb{R}^{N\times 32\times H\times W}$. We then fuse $F_{\text{global}}$ and $F_{\text{geom}}$ via concat and a conv layer to obtain $\mathbf{F}$. We also include the pixel color $\mathbf{I}$, which will be used to provide initialization for Gaussian SH parameter. 
% \vspace{0.5\baselineskip}

\begin{figure}[htbp]
  \centering
  \vspace{-0.8em}
  \includegraphics[width=0.45\textwidth]{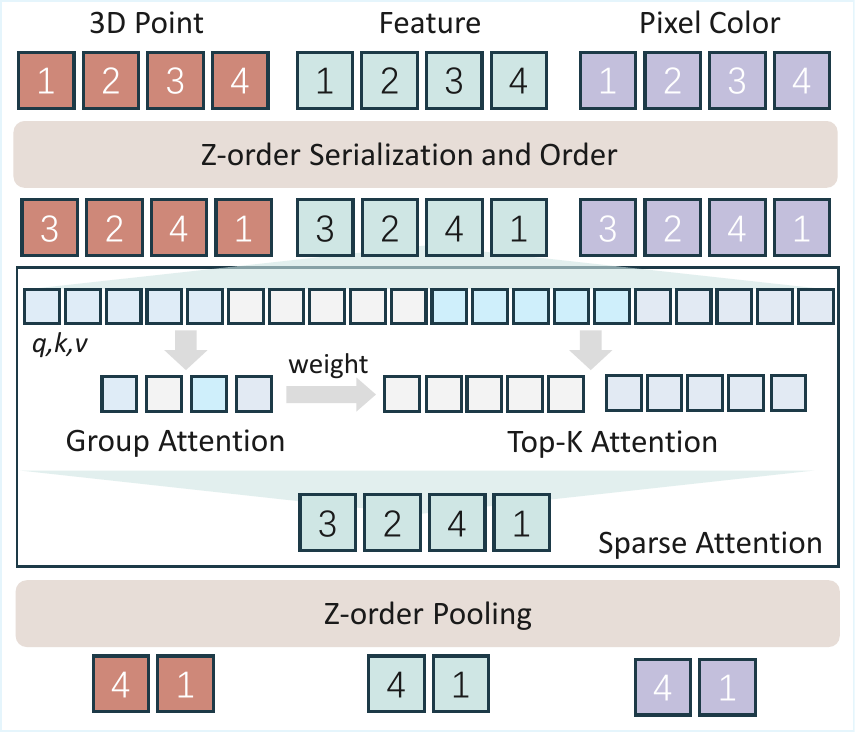}
  \vspace{-0.5em}
  \caption{\textbf{ZFormer Block. }The ZFormer block begins by serializing and ordering the 3D points, features, and pixel colors using Z-order. The serialized data is then passed through a sparse attention mechanism, including group attention, and top-K attention. After the attention steps, Z-order pooling is applied to further aggregate the features.
}
  \label{fig:sa}
  % \vspace{-0.6em}
\end{figure}

\noindent\textbf{ZFormer Blocks}. Given the Gaussian point representation $\mathbf{R}=\left \{\mathbf{P}, \mathbf{F},\mathbf{I}\right \} $, initially, the 3D points $\mathbf{P}$ will be merged along the view dimension to obtain $\mathbf{\hat{P}}\in \mathbb{R}^{(NHW)\times 3}$ and then encoded to Z-order code $\mathbf{Z}\in \mathbb{R}^{(NHW)\times 1}$ via Z-order serialization (Eq.~(\ref{eq:zorder})).
After that, we reorder $\mathbf{R}$ by sorting $\mathbf{Z}$ in ascending order. This procedure arranges the points along the Z-order curve, such that spatially adjacent points in 3D space are also placed in close proximity in the sequence.  
The resulting ordered set $\mathbf{R}=\left \{\mathbf{P}_{\text{sorted}}, \mathbf{F}_{\text{sorted}},\mathbf{I}_{\text{sorted}}\right \} $ preserves spatial locality and provides a structured input representation.

This structured spatial locality motivates us to design a group attention to aggregate sequential blocks of keys or values into block-level representations to capture local information within each group. 
Given the input feature sequence $\mathbf{F}_{\text{sorted}}$, we first project it into query, key, and value embeddings:
\begin{equation}
    \mathbf{Q} = \mathbf{F}_{\text{sorted}}W_Q,\quad \mathbf{K} = \mathbf{F}_{\text{sorted}}W_K,\quad \mathbf{V} = \mathbf{F}_{\text{sorted}}W_V
\end{equation}
where $\mathbf{Q},\mathbf{K},\mathbf{V} \in\mathbb{R}^{(NHW)\times (96/3)}$. The sequence is partitioned into $B=\frac{N\times H\times W}{L} $ non-overlapping blocks of length $L$. We then define a block-wise operator $\mathcal{C}$ that performs average pooling over sequential blocks:
\begin{equation}
    \mathcal{C}(\mathbf{X})_i = \frac{1}{L} \sum_{j=1}^{L} \mathbf{X}_{(i-1)L + j}
\quad \text{for} \quad i = 1, 2, \dots, B
\end{equation}
We then apply this operator to all three projected matrices, yielding the block-wise representations:
\begin{equation}
   \hat{\mathbf{Q}} = \mathcal{C}(\mathbf{Q}), \quad
\hat{\mathbf{K}} = \mathcal{C}(\mathbf{K}), \quad
\hat{\mathbf{V}} = \mathcal{C}(\mathbf{V})
\end{equation}
The group attention is calculated using the scaled dot-product attention:
\begin{equation}
\mathbf{Attn}_{\text{grp}}(\hat{\mathbf{Q}}, \hat{\mathbf{K}}, \hat{\mathbf{V}}) = \text{softmax}\left( \frac{\hat{\mathbf{Q}}\hat{\mathbf{K}}^\top}{\sqrt{d}} \right)\hat{\mathbf{V}}
\label{eq:grp}
\end{equation}
While this group attention significantly reduces computational complexity and captures spatial locality, the averaging operation may dilute fine-grained information within each block. Our solution is to use the top-k attention strategy~\cite{gupta2021memory,vyas2020fast} to select only the top-k most relevant blocks, while reducing attention complexity, where the key is the selection strategy. Inspired by \cite{yuan2025native}, we use the weight $w=\text{softmax}\left( \frac{\hat{\mathbf{Q}}\hat{\mathbf{K}}^\top}{\sqrt{d}} \right)$ from Eq.~(\ref{eq:grp}) to provide differentiable block selection importance scores. Then the top-k attention is formulated as:
\begin{equation}
    \mathbf{Attn}_{\text{sel}} = \text{softmax}\left(\frac{\mathbf{Q}\mathbf{K}_{\text{sel}}^\top}{\sqrt{d}}\right)\mathbf{V}_{\text{sel}}
\end{equation}
Following \cite{yuan2025native}, we then adaptively fuse both attention scores using a gate network $g$:
\begin{equation}
\mathbf{F}_{\text{gate}}=g_1(\mathbf{F}_{\text{sorted}})\odot\mathbf{Attn}_{\text{grp}}+g_2(\mathbf{F}_{\text{sorted}})\odot\mathbf{Attn}_{\text{sel}}
\end{equation}

Currently, we have the Gaussian point representation $\mathbf{R}=\left \{\mathbf{P}_{\text{sorted}}, \mathbf{F}_{\text{gate}},\mathbf{I}_{\text{sorted}}\right \} $. We then introduce a Z-order pooling to reduce the number of points. For the Z-order code $\mathbf{Z}\in \mathbb{R}^{(NHW)\times 1}$, defining the pooling depth as $h$, we have $\mathbf{Z}=\mathbf{Z}\texttt{>>}h$, where $\texttt{>>}$ denotes the bitwise right shift operator. Points with the same Z-order code are grouped into the same cluster. We then conduct an average pooling in each cluster and add another linear projection for the feature part, and obtain the compressed Gaussian point representation: $\mathbf{R}=\left \{\mathbf{P}_{\text{pool}}, \mathbf{F}_{\text{pool}},\mathbf{I}_{\text{pool}}\right \} $, where $\mathbf{P}_{\text{pool}}\in \mathbb{R}^{M\times 3}$ is obtained by inversing Z-order transformation function Eq.~(\ref{eq:zorder}), $\mathbf{F}_{\text{pool}}\in \mathbb{R}^{M\times 96}$, and $\mathbf{I}_{\text{pool}}\in \mathbb{R}^{M\times 3}$. Note $M$ is much smaller than $(NHW)$.
This new Gaussian point representation can be further used for the Z-order block to aggregate points and reduce the number of points.
Suppose there are two Z-order blocks, we will obtain two Gaussian point representations $\mathbf{R}_{L1}$ and $\mathbf{R}_{L2}$.
\vspace{0.5\baselineskip}

\noindent\textbf{Gaussian Head}. We use a two-layer MLP network $\mathcal{F}_{\text{head}}$ to predict Gaussian parameters from various layer Gaussian point representations $\mathbf{R}_{L1}$ and $\mathbf{R}_{L2}$:
\begin{equation}
G_{L1},G_{L2}=\mathcal{F}_{\text{head}}(\mathbf{R}_{L1}),\mathcal{F}_{\text{head}}(\mathbf{R}_{L2})
\end{equation}
Two key points have been considered here: first, we use $\mathbf{P}_{\text{pool}}$ as the base Gaussian centers, and the final Gaussian center is $\mu=\mathbf{P}_{\text{pool}}+\Delta \mu$, where $\Delta \mu$ is predicted by $\mathcal{F}_{\text{head}}$.
Second, we initialize the SH parameter $c$ using the pixel color $\mathbf{I}_{\text{pool}}$ by converting it into SH coefficients.

\subsection{Model Training}
We use the pre-trained depth-anything-v2-small \cite{yang2024depth} $\mathcal{\hat{F}}_{\text{depth}}$ to initialize our depth estimator $\mathcal{F}_{\text{depth}}$. During training, we find that multi-task training with depth estimation favors local details. Thus, rather than fixing the depth estimator, we use $\mathcal{\hat{F}}_{\text{depth}}$ to distill by adding a loss:  
\begin{equation}
    \mathcal{L}_{\text{depth}}=\left | \mathcal{F}_{\text{depth}}\mathbf{I})-\mathcal{\hat{F}}_{\text{depth}}(\mathbf{I}) \right | 
\end{equation}
At the same time, we train the full model with a combination of mean squared error (MSE) and LPIPS~\cite{zhang2018unreasonable} losses between the
rendered and ground-truth image colors:
\begin{equation}
\mathcal{L}_{\text{color}}=\sum_{i=1}^{M}\left [ \text{MSE}(\mathcal{R}(G_{Li},\mathbf{c}) ,\mathbf{I}_{\text{gt}}) + \text{LPIPS}(\mathcal{R}(G_{Li},\mathbf{c}) ,\mathbf{I}_{\text{gt}})\right ] 
\end{equation}
where $M$ is the Z-order block number, $\mathbf{c}$ is the rendered camera, and $\mathcal{R}$ is the Gaussian render. 

\subsection{Inference with Dense Views}

Dense views may be available during inference; however, an excessive number of views can negatively impact inference efficiency. To address this, we propose a Z-order based Maximum Coverage Viewpoint Selection algorithm that effectively reduces redundancy in these dense views, improving efficiency while maintaining performance.

The key of our algorithm is to maximize coverage during view selection.
Given a set of point maps $\mathcal{V}$ from $N$ input views, the algorithm first applies Z-order serialization to encode each point map into a compact point-based representation. The Z-order coverage for each viewpoint is then stored in a max-heap. In each iteration, the algorithm selects the viewpoint offering the largest coverage from the heap and adds it to the set of selected viewpoints $\mathcal{S}$, while updating the covered set $\mathcal{C}$. If no additional viewpoints in the heap contribute sufficient new coverage, the algorithm terminates. By adopting this greedy strategy, the algorithm incrementally selects the most informative viewpoints, thereby reducing redundancy and enhancing computational efficiency. We provide a detailed description of the algorithmic process in Supplementary Algorithm 1.

\begin{table*}[htbp]
\centering
\resizebox{0.9\linewidth}{!}{
\setlength{\tabcolsep}{6pt} 
\renewcommand{\arraystretch}{1.1}
\begin{tabular}{l ccc ccc ccc ccc}
\toprule
\multirow{2}{*}{Method} 
& \multicolumn{3}{c}{2 Views} 
& \multicolumn{3}{c}{4 Views} 
& \multicolumn{3}{c}{8 Views} 
& \multicolumn{3}{c}{12 Views} \\
\cmidrule(lr){2-4} \cmidrule(lr){5-7} \cmidrule(lr){8-10} \cmidrule(lr){11-13}
 & PSNR↑ & SSIM↑ & LPIPS↓ 
 & PSNR↑ & SSIM↑ & LPIPS↓ 
 & PSNR↑ & SSIM↑ & LPIPS↓ 
 & PSNR↑ & SSIM↑ & LPIPS↓ \\
\midrule
\multicolumn{13}{c}{RealEstate10K, 360$\times$640} \\
\midrule
3DGS~\cite{kerbl20233d}          & 16.80 & 0.591 & 0.351 & 21.29 & 0.753 & 0.279 & 23.02 & 0.783 & 0.257 & 26.73 & 0.871 & 0.144 \\
MipSplatting~\cite{yu2024mip}  & 18.09 & 0.703 & 0.310 & 23.46 & 0.801 & 0.214 & 24.87 & 0.812 & 0.198 & 27.34 & 0.891 & \cellcolor{yellow!50}0.117 \\
DepthSplat~\cite{xu2025depthsplat}     & 26.03 & \cellcolor{yellow!50}0.873 & \cellcolor{yellow!50}0.158 & 26.13 & 0.872 & 0.157 & 26.17 & 0.876 & 0.152 & 26.33 & 0.880 & 0.143 \\
AnySplat~\cite{jiang2025anysplat}       & 22.55 & 0.757 & 0.229 & 25.86 & 0.824 & 0.152 & 26.71 & \cellcolor{yellow!50}0.886 & \cellcolor{yellow!50}0.131 & 26.94 & 0.892 & 0.122 \\
Ours\#L1       & \cellcolor{pink}\textbf{26.43} & \cellcolor{pink}\textbf{0.873} & \cellcolor{pink}\textbf{0.147} & \cellcolor{pink}\textbf{26.68} & \cellcolor{pink}\textbf{0.891} & \cellcolor{pink}\textbf{0.142} & \cellcolor{pink}\textbf{27.25} & \cellcolor{pink}\textbf{0.897} & \cellcolor{pink}\textbf{0.123} & \cellcolor{pink}\textbf{28.56} & \cellcolor{pink}\textbf{0.901} & \cellcolor{pink}\textbf{0.110} \\
Ours\#L2       & \cellcolor{yellow!50}26.42 & 0870 & 0.159 & \cellcolor{yellow!50}26.59 & \cellcolor{yellow!50}0.884 & \cellcolor{yellow!50}0.144 & \cellcolor{yellow!50}26.78 & 0.884 & 0.132 & \cellcolor{yellow!50}28.12 & \cellcolor{yellow!50}0.898 & 0.120 \\
\midrule
\multicolumn{13}{c}{DL3DV 256$\times$448} \\
\midrule
3DGS~\cite{kerbl20233d}          & 18.79 & 0.571 & 0.311 & 19.51  & 0.656 & 0.288 & 21.70 & 0.736 & 0.200 & 23.63 & 0.810 & 0.159 \\
MipSplatting~\cite{yu2024mip}  & 19.10 & 0.657 & 0.288 & 20.28 & 0.649 & 0.246 & 22.47 & 0.713 & 0.235 & 23.79 & 0.819 & 0.137\\
DepthSplat~\cite{xu2025depthsplat}     & 20.23 & 0.679 & 0.268 & 23.08 & 0.760 & 0.183 & 24.15 & 0.810 & \cellcolor{yellow!50}0.141 & 25.74 & 0.844 & \cellcolor{yellow!50}0.131 \\
AnySplat~\cite{jiang2025anysplat}       & 20.65 & 0.669 & 0.283 & 22.85 & 0.746 & 0.234 & 24.38 & 0.823 & 0.167 & 24.90 & 0.856 & 0.136 \\
Ours\#L1       & \cellcolor{pink}\textbf{23.06} & \cellcolor{pink}\textbf{0.774} & \cellcolor{pink}\textbf{0.201} & \cellcolor{pink}\textbf{24.95} & \cellcolor{pink}\textbf{0.845} & \cellcolor{pink}\textbf{0.143} & \cellcolor{pink}\textbf{26.22} & \cellcolor{pink}\textbf{0.855} & \cellcolor{pink}\textbf{0.137} & \cellcolor{pink}\textbf{27.09} & \cellcolor{pink}\textbf{0.892} & \cellcolor{pink}\textbf{0.124} \\
Ours\#L2       & \cellcolor{yellow!50}22.99 & \cellcolor{yellow!50}0.770 & \cellcolor{yellow!50}0.206 & \cellcolor{yellow!50}24.55 & \cellcolor{yellow!50}0.830 & \cellcolor{yellow!50}0.153 & \cellcolor{yellow!50}25.53 & \cellcolor{yellow!50}0.846 & 0.142 & \cellcolor{yellow!50}26.62 & \cellcolor{yellow!50}0.876 & \cellcolor{yellow!50}0.131 \\
\bottomrule
\end{tabular}
}
\vspace{-0.6em}
\caption{\textbf{Quantitative Comparison with Varying Numbers of Views.} We highlight the best results in red and the second-best in yellow.}
\label{tab:comparison-fixview}
\vspace{-1em}
\end{table*}

\section{Experiment}
\label{sec:exp}

\noindent\textbf{Datasets.} We evaluate our method using three large-scale datasets: the $360\times640$ resolution RealEstate10K~\cite{zhou2018stereo}, the $256\times448$ resolution DL3DV~\cite{ling2024dl3dv}, and the $256\times256$ resolution ACID~\cite{liu2021infinite}. The training and testing splits follow the protocol of DepthSplat~\cite{xu2025depthsplat}. For RealEstate10K and DL3DV, we use the provided estimated camera intrinsic and extrinsic parameters, while for ACID, we derive the camera information using VGGT~\cite{wang2025vggt}. The main experiments are conducted on the RealEstate10K and DL3DV datasets. ACID is not used for training; instead, it is employed exclusively for cross-dataset evaluation.
\vspace{0.5em}

\noindent\textbf{Implementation Details. }
In our framework, we utilize two ZFormer blocks. The block size for group attention is set to 32, while for top-k attention, we select half of the total number of blocks. To accelerate the attention computation, we use FlashAttention~\cite{shah2024flashattention}. The Z-order pooling depth is set to 2. Our method is implemented in PyTorch, and we optimize the model using the AdamW~\cite{loshchilovdecoupled} optimizer with a cosine learning rate schedule.
We use a lower learning rate of $2 \times 10^{-6}$ for the depth estimation branch, while the ZFormer blocks and the Gaussian head are trained with a learning rate of $2 \times 10^{-4}$. The model is trained on $8 \times$ A100 GPUs for 100K iterations with a batch size of 2, taking approximately 2 days.
For quantitative results, we report standard image quality metrics, including pixel-level PSNR, patch-level SSIM~\cite{wang2004image}, and feature-level LPIPS~\cite{zhang2018unreasonable}.

\subsection{Comparisons}

\begin{figure*}[htbp]
  \centering
  \includegraphics[width=0.98\textwidth]{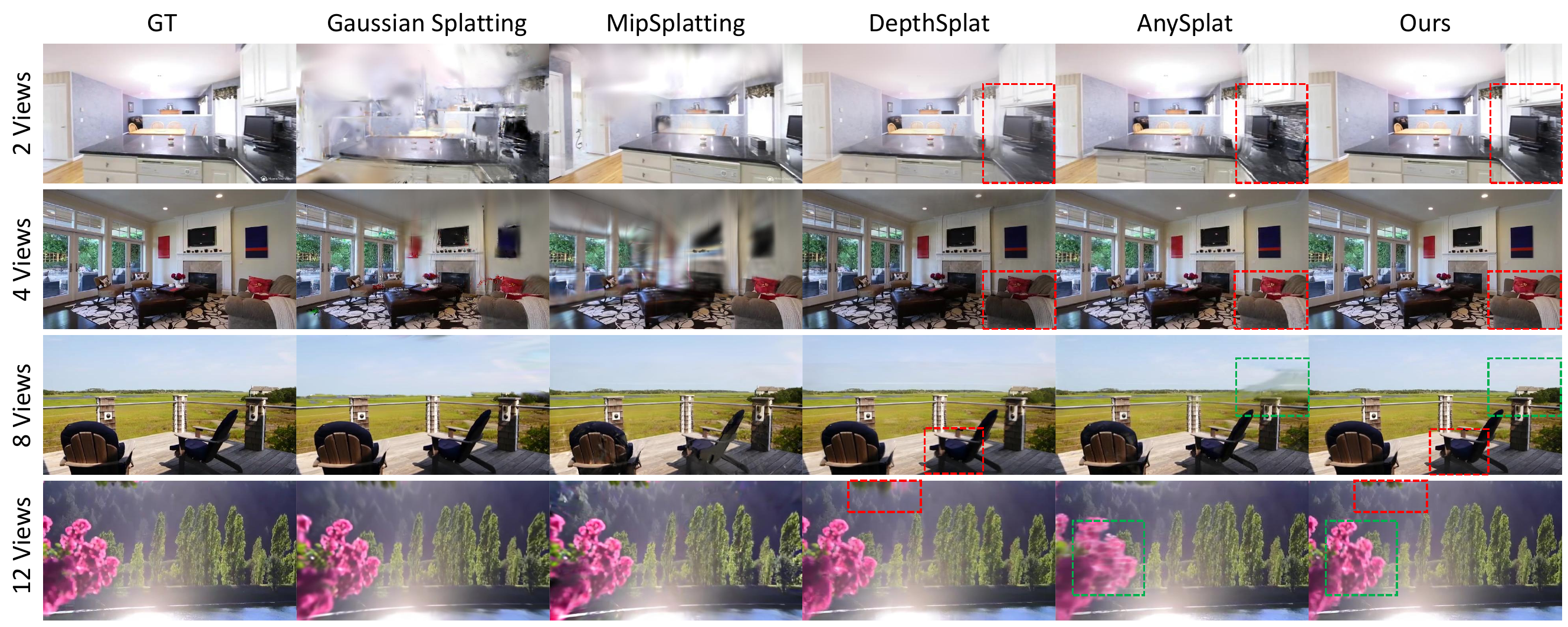}
  \vspace{-0.6em}
  \caption{\textbf{Visual Comparison.} Our
method achieves better performance in capturing sharp edges and intricate details.}
  \label{fig:cmp}
  \vspace{-1em}
\end{figure*}

We compare our method with two classical optimization-based methods, 3DGS~\cite{kerbl20233d} and MipSplatting~\cite{yu2024mip}, as well as two representative feed-forward methods, DepthSplat~\cite{xu2025depthsplat} and AnySplat~\cite{jiang2025anysplat}. For all methods, we use the same input views with an identical resolution for comparison. Additionally, for the feed-forward methods, we use the same training data and the official training script during the training process.

\noindent\textbf{Comparisons on Fixed View Input.} We compare the results for different numbers of input views (2, 4, 8, and 12) in Tab.~\ref{tab:comparison-fixview}. Our method with a layer-one Z-order block (Ours\#L1) consistently outperforms related works across all metrics and for all view counts. In most cases, our method with two Z-order block compressions (Ours\#L2) outperforms the comparison methods, achieving better results across various metrics and input configurations.
Our improvements are particularly significant with fewer input views, especially when using only two views. This demonstrates that our model is more robust to sparse views.
We also provide a visual comparison in Fig.~\ref{fig:cmp}, where our method demonstrates better performance, surpassing the baselines in capturing sharp edges and intricate details.
\vspace{0.5em}

\noindent\textbf{Comparisons on Variable Input.}
In Tab.~\ref{tab:comparison-merged}, we evaluate the performance of various feed-forward methods across different view inputs, ranging from 2 to 12 views. All methods are initially trained on the RealEstate10K dataset for 100K iterations and then fine-tuned on the DL3DV dataset for an additional 50K iterations, with view inputs varying from 2 to 12. Both Ours\#L1 and Ours\#L2 outperform the baseline methods, DepthSplat and AnySplat, across all three metrics. Ours\#L1 achieves the best overall performance, while Ours\#L2 slightly trails behind, demonstrating the superior image quality, structural consistency, and perceptual fidelity of our approach.
\vspace{0.5em}

\noindent\textbf{Cross-Dataset Comparisons.}
We assess the cross-dataset generalization capability by testing the RealEstate10K and DL3DV pre-trained models on the ACID dataset in Tab.~\ref{tab:cross_dataset_eval}. Ours\#L1 achieves the best performance in both RealEstate10K$\rightarrow$ACID and DL3DV$\rightarrow$ACID evaluations. Ours\#L2 also shows strong results, closely trailing Ours\#L1 across all metrics. These findings highlight the superior generalization ability of our methods across datasets. 
\vspace{0.5em}

\noindent\textbf{Inference Time and Number of Gaussians Comparisons.} In Tab.~\ref{tab:runtime_views_res}, our methods, Ours\#L1 and Ours\#L2, significantly outperform the baseline methods in both inference time and the number of Gaussian primitives. Notably, our method is approximately 1,000 times faster than 3D-GS and Mip-Splatting, with a 3× reduction in Gaussian primitives for 2 views. Additionally, our approach achieves a 2- to 3-fold reduction in Gaussian primitives compared to DepthSplat and AnySplat.

\subsection{Ablation Study}

\begin{figure*}[htbp]
  \centering
  \includegraphics[width=0.98\textwidth]{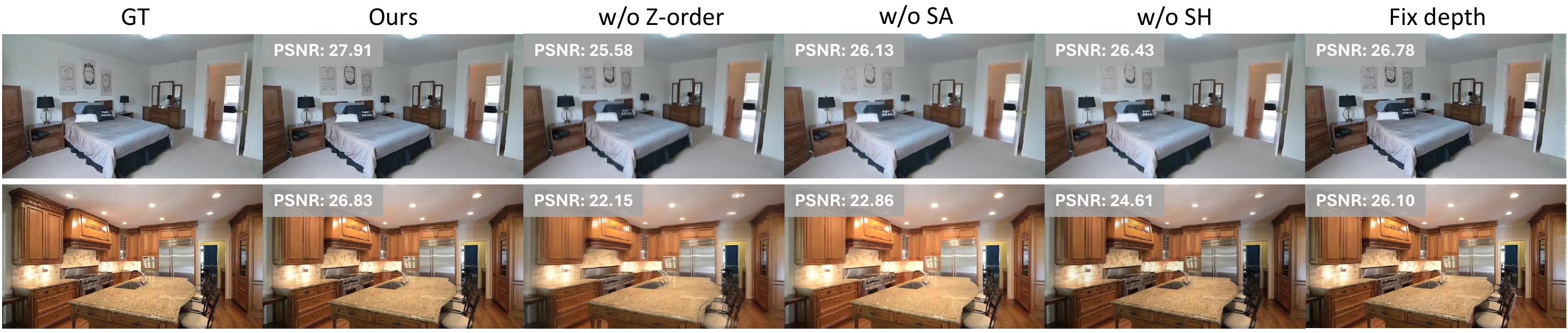}
  \vspace{-0.6em}
  \caption{\textbf{Visual Ablation Study.} Results of ablations on different components of our method.}
  \label{fig:abl}
  \vspace{-1em}
\end{figure*}

\begin{figure}[htbp]
  \centering
  \includegraphics[width=0.40\textwidth]{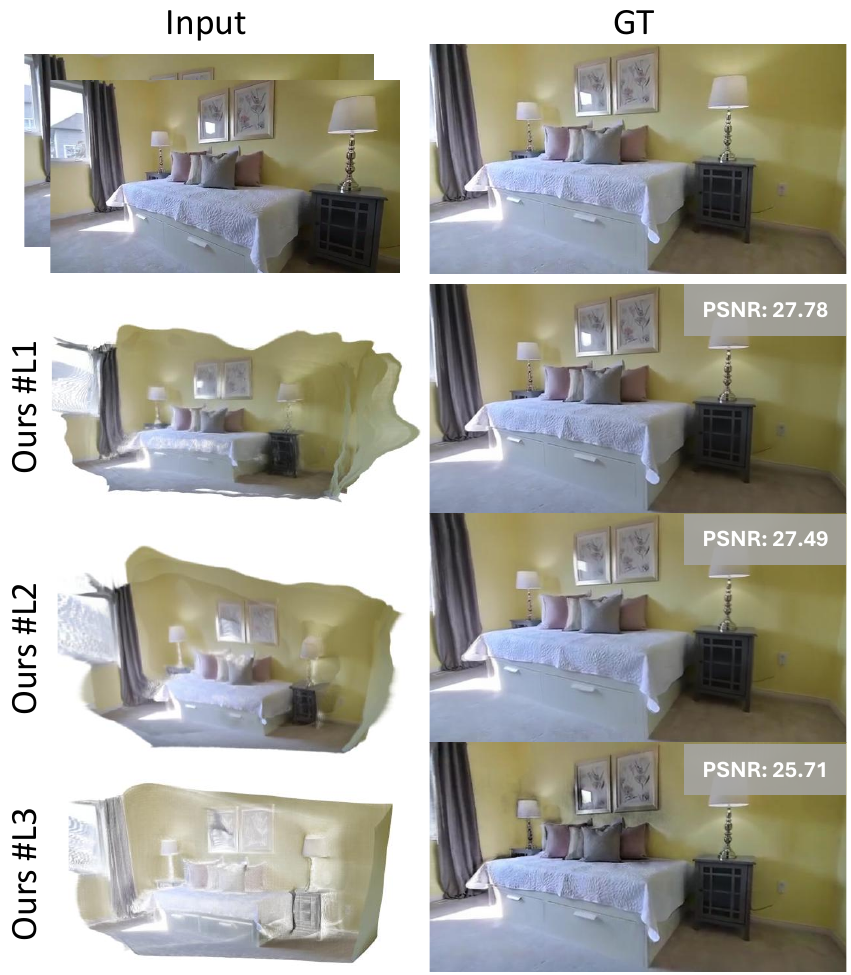}
  \vspace{-0.6em}
  \caption{\textbf{Ablation Study on Layer Selection.} We use two Z-order blocks in our framework to prevent degradation while achieving a lower number of GS primitives.
}
  \label{fig:layer}
  \vspace{-1em}
\end{figure}

We conduct six ablation experiments using 12 input views in Tab.~\ref{tab:ablation}, Tab.~\ref{tab:selection_comparison}, Fig.~\ref{fig:abl}, and Fig.~\ref{fig:layer} to evaluate the impact of various design choices on the performance of our model, including model structure, training strategy, Z-order layer selection, and view selection during inference.

\noindent\textbf{Model Design.} In Tab.~\ref{tab:ablation}, replacing the ZFormer blocks with convolutional layers (Ours w/o Z-order) leads to a noticeable decrease in all metrics (resulting in the worst performance), suggesting that the ZFormer block is essential in our framework. Using Z-order while replacing sparse attention with a full attention layer (Ours w/o SA) also results in a clear performance drop compared to our full model, indicating that sparse attention plays a crucial role in our model design. These observations can also be confirmed in Fig.~\ref{fig:abl}, where the results tend to be coarse both globally and locally without ZFormer or sparse attention.
\vspace{0.5em}

\noindent\textbf{Training Strategy.}
During training, we found it important to initialize SH parameters using the pixel color from the input views, as this enhances training stability and leads to better convergence (refer to Ours w/o SH in Tab.\ref{tab:ablation} and Fig.\ref{fig:abl}).
Moreover, by comparing our method without multi-task depth training (Ours-Fix-Depth) to the full model in Tab.~\ref{tab:ablation} and Fig.~\ref{fig:abl}, we find that it is important not to fix depth estimation during training. This is because 2D image-based depth estimation is not always accurate, and joint training allows the model to compensate for such inaccuracies.
\vspace{0.5em}

\noindent\textbf{Z-order Layer Selection.}
We explain the selection of two Z-order layers in Fig.~\ref{fig:layer}. We select two Z-order layers in our framework because they strike a balance between preventing degradation and maintaining a low number of GS primitives (also see Tab.~\ref{tab:runtime_views_res}). These two layers help preserve fine details and overall structure, as evidenced by the degradation observed when more than two layers are used. The optimal performance achieved with two layers reflects a careful trade-off between quality and efficiency.
\vspace{0.5em}

\noindent\textbf{Inference View Selection.}
In Tab.~\ref{tab:selection_comparison},  we investigate the impact of different view selection strategies on the performance and efficiency of our model.
Our Z-order selection consistently outperforms random selection in terms of image quality. It also provides a more efficient alternative to using all views, with slightly longer inference times compared to random selection. This suggests our Z-order selection is an effective method for balancing both performance and computational efficiency.

\begin{table}[t]
\centering
\resizebox{0.68\linewidth}{!}{
\setlength{\tabcolsep}{6pt} % Slightly increase column spacing
\renewcommand{\arraystretch}{1.1}
\begin{tabular}{l ccc ccc ccc}
\toprule
\multirow{2}{*}{Method} 
& \multicolumn{3}{c}{2-12 Views} \\
\cmidrule(lr){2-4}
 & PSNR↑ & SSIM↑ & LPIPS↓ \\
\midrule
DepthSplat~\cite{xu2025depthsplat}     & 26.11 &  0.871 &  0.151 \\
AnySplat~\cite{jiang2025anysplat}       & 26.54 & 0.875 &  0.133 \\
Ours\#L1       & \cellcolor{pink}\textbf{28.07} & \cellcolor{pink}\textbf{0.890} & \cellcolor{pink}\textbf{0.125} \\
Ours\#L2       & \cellcolor{yellow!50}27.62 & \cellcolor{yellow!50}0.881 & \cellcolor{yellow!50}0.129 \\
\bottomrule
\end{tabular}
}
\caption{\textbf{Comparison on Variable Input.} We train and evaluate the performance across
various view inputs (2 to 12 views).}
\label{tab:comparison-merged}
\end{table}

\begin{table}[t]
\centering
\setlength{\tabcolsep}{4pt} 
\renewcommand{\arraystretch}{1.1} 
\resizebox{\linewidth}{!}{
\begin{tabular}{l ccc ccc}
\toprule
\multirow{2}{*}{Method} 
& \multicolumn{3}{c}{RealEstate10K $\rightarrow$ ACID} 
& \multicolumn{3}{c}{DL3DV $\rightarrow$ ACID} \\
\cmidrule(lr){2-4} \cmidrule(lr){5-7}
& PSNR↑ & SSIM↑ & LPIPS↓ 
& PSNR↑ & SSIM↑ & LPIPS↓ \\
\midrule
DepthSplat~\cite{xu2025depthsplat} & 26.05 & 0.810 & \cellcolor{yellow!50}0.181 & 25.58 & 0.796 & 0.203 \\
AnySplat~\cite{jiang2025anysplat}   & 22.71 & 0.685 & 0.298 & 23.64 & 0.737 & 0.242 \\
Ours\#L1   & \cellcolor{pink}\textbf{27.56} & \cellcolor{pink}\textbf{0.853} & \cellcolor{pink}\textbf{0.172} 
           & \cellcolor{pink}\textbf{27.34} & \cellcolor{pink}\textbf{0.845} & \cellcolor{pink}\textbf{0.169} \\
Ours\#L2   & \cellcolor{yellow!50}27.01 & \cellcolor{yellow!50}0.824 & 0.183 
           & \cellcolor{yellow!50}26.95 & \cellcolor{yellow!50}0.831 & \cellcolor{yellow!50}0.187 \\
\bottomrule
\end{tabular}
}
\vspace{-0.6em}
\caption{\textbf{Cross-Dataset Evaluation.} We assess the cross-dataset generalization capability by testing the RealEstate10K and DL3DV pre-trained models on the ACID dataset.}
\label{tab:cross_dataset_eval}
\vspace{-1em}
\end{table}

\begin{table}[t]
\centering
\setlength{\tabcolsep}{4pt} 
\renewcommand{\arraystretch}{1.1} 
\resizebox{0.35\textwidth}{!}{ 
\begin{tabular}{l c c c c c}
\toprule
\multirow{2}{*}{Method} &
\multicolumn{2}{c}{2 Views} & 
\multicolumn{2}{c}{12 Views} \\
\cmidrule(lr){2-3} \cmidrule(lr){4-5}
& {Opt./Inf.} & {\#GS} & {Opt./Inf.} & {\#GS} \\
\midrule
3DGS~\cite{kerbl20233d}        & 2m15s  & 6.27  & 8m21s  & 8.51   \\
MipSplatting~\cite{yu2024mip} & 1m18s  & 4.39  & 5m14s  & \cellcolor{yellow!50}\textbf{8.14}   \\
DepthSplat~\cite{xu2025depthsplat}   & 0.142s & 4.61  & 0.384s & 27.6   \\
AnySplat~\cite{jiang2025anysplat}     & 0.692s & 3.53  & 1.212s & 13.2   \\
Ours\#L1     & \cellcolor{pink}\textbf{0.123s}     &  \cellcolor{yellow!50}2.85   &  \cellcolor{pink}\textbf{0.337s}    & 17.8   \\
Ours\#L2     & \cellcolor{yellow!50}0.135s     & \cellcolor{pink}\textbf{1.42}    & \cellcolor{yellow!50}0.355s & \cellcolor{pink}8.05   \\
\bottomrule
\end{tabular}}
\vspace{-0.6em}
\caption{\textbf{Comparison of Runtime and the Number of Gaussian Primitives. } Values are reported in milliseconds (ms) for 2 and 12 views, with a resolution of 360×640. \#GS denotes the number of Gaussian primitives ($\times 10^5$). }
\label{tab:runtime_views_res}
\vspace{-1em}
\end{table}

\begin{table}[t]
\centering
\resizebox{0.3\textwidth}{!}{
\setlength{\tabcolsep}{2pt} 
\renewcommand{\arraystretch}{1.1}
\begin{tabular}{lcccc}
\toprule
Method               & PSNR↑ & SSIM↑ & LPIPS↓   \\
\midrule
Ours w/o Z-order     &  24.86    &  0.794    &   0.225                      \\
Ours w/o SA &   26.79   &   0.847   &  0.174                  \\
Ours w/o SH  &  27.81    &   0.874   &   0.129                      \\
Ours-Fix-Depth  &   28.15   &   0.893   &    0.119                     \\
Ours                 &   \textbf{28.56}   &  \textbf{ 0.901}   &   \textbf{0.110}   \\
\bottomrule
\end{tabular}}
\vspace{-0.6em}
\caption{\textbf{Ablation Study.} We have conducted ablations on different components of our method.}
\label{tab:ablation}
\vspace{-1em}
\end{table}

\begin{table}[t]
\centering
\resizebox{0.4\textwidth}{!}{
\setlength{\tabcolsep}{2.4pt} % Adjust column spacing
\renewcommand{\arraystretch}{1.2} % Adjust row height
\begin{tabular}{ccccc c}
\toprule
Inp. Views & Selection & PSNR↑ & SSIM↑ & LPIPS↓ & Time(s)↓ \\
\midrule
\multirow{3}{*}{24}
& NA. & \textbf{28.91} & \textbf{0.906} & \textbf{0.102} & 0.622 \\ \cline{2-6}
 & RS. 16 & 27.97 & 0.891 & 0.113 & \textbf{0.417} \\ \cline{2-6}
 & ZS. 16 & 28.73 & 0.903 & 0.108 & 0.448 \\ \hline
\multirow{3}{*}{16}
& NA. & \textbf{28.67} & \textbf{0.901} & \textbf{0.110} & 0.417 \\ \cline{2-6}
 & RS. 8 & 27.06 & 0.876 & 0.125  & \textbf{0.255} \\ \cline{2-6}
 & ZS. 8 & 28.07 & 0.898 & 0.116  & 0.272 \\
\bottomrule
\end{tabular}}
\vspace{-0.6em}
\caption{\textbf{Ablation Study with Different Selection Strategies.} NA. indicates that all views will be used during inference, RS. refers to random selection, while ZS. denotes our Z-order-based view selection method. The input resolution is 360$\times$640.}
\label{tab:selection_comparison}
\vspace{-1em}
\end{table}

\section{Conclusion}
\label{sec:con}

In this work, we have proposed a Z-order transformer framework for efficient 3DGS and novel view synthesis. By employing a Z-order strategy, we arrange Gaussian primitives into a spatial sequence, enabling effective context modeling and efficient prediction in a single forward pass.
Furthermore, we propose a sparse attention mechanism to reduce the complexity of processing large-scale Gaussian data, significantly enhancing both the efficiency and quality of the reconstruction.
Through extensive experiments, we have demonstrated that our method outperforms existing state-of-the-art approaches. Notably, our method achieves faster inference times while requiring fewer Gaussian primitives for high-quality novel view synthesis.
In summary, our Z-order transformer framework advances 3DGS, offering an efficient solution for real-time, high-quality novel view synthesis with fewer Gaussian primitives.


% \section*{Acknowledgements}
% This work is supported in part by the Research Grants Council (RGC) of the Hong Kong SAR under the General Research Fund (17203023), the Collaborative Research Fund (C5052-23G), and the NSFC/RGC Collaborative Research Scheme (CRS-HKU703/24). The research work described in this paper was conducted in the JC STEM Lab of Multimedia and Machine Learning funded by The Hong Kong Jockey Club Charities Trust.

{
    \small
    \bibliographystyle{ieeenat_fullname}
    \bibliography{main}
}

\clearpage
\twocolumn[{
\begin{center}
    \Large \textbf{Supplemental Material} \\[1em]
\end{center}
}]

\setcounter{section}{0}
\renewcommand{\thesection}{\arabic{section}}

\section{Z-order Based Maximum Coverage Viewpoint Selection Algorithm}

\begin{algorithm}
\caption{Z-order Based Maximum Coverage Viewpoint Selection with Max-Heap}
\label{alg:viewpoint-selection-maxheap}
\begin{algorithmic}[1]
\Require 
Viewpoint set $\mathcal{V} = \{V_1, V_2, \ldots, V_N\}$,
Number of viewpoints to select $M$,
Grid size $\delta$
\Ensure Selected viewpoint indices $\mathcal{S}$, final coverage $C$

\State $\mathcal{C} \gets \emptyset$ \Comment{Set of covered grid cells}
\State $\mathcal{S} \gets \emptyset$ \Comment{Set of selected viewpoints}
\State $\mathcal{H} \gets \emptyset$ \Comment{Max-heap (coverage, viewpoint index)}

\For{$i = 1$ \textbf{to} $N$}
    \State $P_i \gets \text{Get point cloud for viewpoint } V_i$
    \State $S_i \gets \{\mathcal{Z}(\phi(p/\delta)) \mid p \in P_i\}$ \Comment{Z-order serialization with grid size $\delta$}
    \State $\text{push } (|S_i|, i) \text{ into } \mathcal{H}$ \Comment{Push initial coverage into heap}
\EndFor

\For{$k = 1$ \textbf{to} $M$}
    \If{$\mathcal{H} = \emptyset$}
        \State \textbf{break}
    \EndIf
    
    \State $\text{found} \gets \text{False}$
    \While{$\mathcal{H} \neq \emptyset$ \textbf{and not} found}
        \State $(c, j) \gets \text{pop from } \mathcal{H}$ \Comment{Get max coverage candidate}
        \State $\Delta \gets S_j \setminus \mathcal{C}$ \Comment{Find new covered grid cells for viewpoint $j$}
        
        \If{$|\Delta| > 0$}
            \State $\mathcal{C} \gets \mathcal{C} \cup \Delta$ \Comment{Add new covered grid cells}
            \State $\mathcal{S} \gets \mathcal{S} \cup \{j\}$ \Comment{Add viewpoint $j$ to selected viewpoints}
            \State $\text{found} \gets \text{True}$
            \State \textbf{break}
        \EndIf
    \EndWhile
    
    \If{\textbf{not} found}
        \State \textbf{break} \Comment{No more coverage can be added}
    \EndIf
    
    \State $\mathcal{H}_{\text{new}} \gets \emptyset$
    \While{$\mathcal{H} \neq \emptyset$}
        \State $(c, j) \gets \text{pop from } \mathcal{H}$
        \State $\Delta_j \gets S_j \setminus \mathcal{C}$ \Comment{Recalculate new coverage for viewpoint $j$}
        \If{$|\Delta_j| > 0$}
            \State $\text{push } (|\Delta_j|, j) \text{ into } \mathcal{H}_{\text{new}}$
        \EndIf
    \EndWhile
    \State $\mathcal{H} \gets \mathcal{H}_{\text{new}}$
\EndFor

\State \Return $\mathcal{S}, |\mathcal{C}|$
\end{algorithmic}
\end{algorithm}

In the main paper, we provide a high-level introduction to the proposed Z-order based Maximum Coverage Viewpoint Selection algorithm. Here, we include the detailed description of this algorithm, as shown in Algorithm~\ref{alg:viewpoint-selection-maxheap}. 
The proposed algorithm aims to select at most $M$ viewpoints from a dense candidate set $\mathcal{V} = \{V_1, \ldots, V_N\}$ such that their union maximally covers the underlying 3D scene while avoiding redundant views.
For each candidate viewpoint $V_i$, we first obtain its associated point cloud $P_i$ and discretize the 3D space into a regular grid with a cell size $\delta$. 
Each point $p \in P_i$ is mapped to its grid cell and then serialized into a one-dimensional key using a Z-order encoding $\mathcal{Z}(\phi(p/\delta))$, which preserves spatial locality, where $\phi$ denotes the quantization from continuous 3D coordinates to integer grid indices, implemented as component-wise floor $\phi(p/\delta)=\lfloor p/\delta \rfloor$.
The resulting set of encoded grid cells for viewpoint $V_i$ is denoted as $S_i$, and its cardinality $|S_i|$ reflects the raw coverage of that viewpoint. 
We then initialize a max-heap $\mathcal{H}$ with tuples $(|S_i|, i)$ for all viewpoints, so that the viewpoint with the largest coverage can be efficiently retrieved. The algorithm proceeds in a greedy fashion for at most $M$ iterations. At each iteration, we repeatedly pop the current best candidate $(c, j)$ from $\mathcal{H}$ and compute its marginal contribution $\Delta = S_j \setminus \mathcal{C}$ with respect to the already covered set $\mathcal{C}$. If $|\Delta| > 0$, viewpoint $V_j$ is accepted: we update the global coverage $\mathcal{C} \leftarrow \mathcal{C} \cup \Delta$ and append $j$ to the selected viewpoint index set $\mathcal{S}$. If the heap is exhausted without finding a viewpoint that adds new coverage, the selection process terminates early, indicating that remaining viewpoints are completely redundant. After each successful selection, we rebuild the heap by recomputing, for every remaining candidate viewpoint $V_j$, its updated marginal coverage $\Delta_j = S_j \setminus \mathcal{C}$ and pushing only those with $|\Delta_j| > 0$ back into a new heap $\mathcal{H}_{\text{new}}$. Overall, this Z-order based greedy scheme incrementally selects the most informative viewpoints under a discretized spatial coverage criterion, effectively eliminating redundant views and improving inference efficiency while avoiding significant performance degradation.

\begin{table}[t]
\renewcommand{\thetable}{S\arabic{table}}
\centering
\resizebox{0.44\textwidth}{!}{
\setlength{\tabcolsep}{2.4pt} % Adjust column spacing
\renewcommand{\arraystretch}{1.2} % Adjust row height
\begin{tabular}{ccccc c}
\toprule
Inp. Views & Selection & PSNR↑ & SSIM↑ & LPIPS↓ & Time(s)↓ \\
\midrule
\multirow{3}{*}{64}
& NA. & \textbf{29.44} & \textbf{0.911} & \textbf{0.098} & 1.891 \\ \cline{2-6}
 & RS. 16 & 28.50 & 0.897 & 0.118 & \textbf{0.421} \\ \cline{2-6}
 & ZS. 16 & 29.13 & 0.908 & 0.106 & 0.498 \\ \hline
\multirow{3}{*}{24}
& NA. & \textbf{28.91} & \textbf{0.906} & \textbf{0.102} & 0.622 \\ \cline{2-6}
 & RS. 16 & 27.97 & 0.891 & 0.113 & \textbf{0.417} \\ \cline{2-6}
 & ZS. 16 & 28.73 & 0.903 & 0.108 & 0.448 \\ \hline
\multirow{3}{*}{16}
& NA. & \textbf{28.67} & \textbf{0.901} & \textbf{0.110} & 0.417 \\ \cline{2-6}
 & RS. 8 & 27.06 & 0.876 & 0.125  & \textbf{0.255} \\ \cline{2-6}
 & ZS. 8 & 28.07 & 0.898 & 0.116  & 0.272 \\
\bottomrule
\end{tabular}}
% \vspace{-0.6em}
\caption{\textbf{Ablation Study with Different Selection Strategies.} NA. indicates that all views will be used during inference, RS. refers to random selection, while ZS. denotes our Z-order-based view selection method. The input resolution is 360$\times$640.}
\label{tab:selection_comparison_appendix}
% \vspace{-1em}
\end{table}

In Tab.~6 of the main paper, we investigate the impact of different view selection strategies. Here, we further include a denser input setting to extend Tab.~6 and evaluate performance, as shown in Tab.~\ref{tab:selection_comparison_appendix}-top.
For the denser 64-view input setting, our method still performs well, exhibiting only a minor performance drop while removing redundant views. The efficiency improvement is also more pronounced in this case, as we discard half of the input views.

\section{More Comparisons with Related Works}

While our comparison baseline, AnySplat~\cite{jiang2025anysplat}, has already been shown to outperform NoPoSplat~\cite{ye2025nopose} and FLARE~\cite{zhang2025flare}, and DepthSplat~\cite{xu2025depthsplat} has been shown to surpass MVSplat~\cite{chen2024mvsplat} and PixelSplat~\cite{pixelsplat}, we have also included direct comparisons with these methods, as well as MonoSplat~\cite{liu2025monosplat}, to further validate the effectiveness of our approach.
All methods are trained and evaluated on the RealEstate10K dataset~\cite{zhou2018stereo} using the same training and testing split at a resolution of $256\times256$ with 2 input views.
As summarized in Table~\ref{tab:comparison_appendix}, our method consistently delivers better reconstruction quality (higher PSNR and SSIM, lower LPIPS), demonstrating that our design provides clear advantages over these related approaches.

\begin{table}[t]
\renewcommand{\thetable}{S\arabic{table}}
\centering
\begin{tabular}{lccc}
\toprule
Method & PSNR$\uparrow$ & SSIM$\uparrow$ & LPIPS$\downarrow$ \\
\midrule
NoPoSplat~\cite{ye2025nopose} & 27.41 & 0.884 & 0.116 \\
FLARE~\cite{zhang2025flare}     & 23.78 & 0.801 & 0.191 \\
MonoSplat~\cite{liu2025monosplat}     & 26.68 & 0.875 & 0.123 \\
MVSplat~\cite{chen2024mvsplat}     & 26.39  & 0.869 &  0.128 \\
pixelSplat~\cite{pixelsplat}     & 25.89 & 0.858 & 0.142 \\
Ours      & \textbf{27.89} & \textbf{0.892} & \textbf{0.110} \\
\bottomrule
\end{tabular}
\caption{\textbf{More Comparisons. }We present additional comparison results on the RealEstate10K dataset at a resolution of 256$\times$256 with 2 input views.}
\label{tab:comparison_appendix}
\end{table}

\section{Ablation Studies of Sparse Attention}

To quantify the contribution of each component in our sparse attention module, we perform ablation experiments on the RealEstate10K dataset with 12 input views under the same training and evaluation settings as in the main paper.
As reported in Table~\ref{tab:ablation_appendix}, 
disabling the selection mechanism (``Ours w/o sel'') degrades performance, demonstrating that adaptively selecting informative tokens is important for effective sparse attention.
When we remove the entire sparse attention module (``Ours w/o SA''), the performance degrades the most, confirming that group attention and selective attention jointly contribute to the overall gain. The complete model achieves the best reconstruction quality, with the highest PSNR/SSIM and the lowest LPIPS, demonstrating the effectiveness of our sparse attention design.
Note that we do not perform an ablation of the group attention removal, as the selection attention depends on it.

\begin{table}[t]
\renewcommand{\thetable}{S\arabic{table}}
\centering
\resizebox{0.35\textwidth}{!}{
\setlength{\tabcolsep}{2pt} % 调整列间距
\renewcommand{\arraystretch}{1.1} % 调整行高
\begin{tabular}{lcccc}
\toprule
Method               & PSNR↑ & SSIM↑ & LPIPS↓   \\
\midrule
% Ours w/o grp     &  26.86    &  0.794    &   0.225                      \\
Ours w/o sel &   26.79   &   0.847   &  0.174                  \\
Ours w/o SA &   26.79   &   0.847   &  0.174                  \\
Ours                 &   \textbf{28.56}   &  \textbf{ 0.901}   &   \textbf{0.110}   \\
\bottomrule
\end{tabular}}
% \vspace{-0.6em}
\caption{\textbf{Ablation Studies of Sparse Attention.} We conduct ablation experiments on different components of our sparse attention module.
Experiments are conducted on the 
RealEstate10K dataset at a resolution of 360$\times$640 with 12 input views.
Note that we do not perform an ablation of the group attention removal, as the selection attention depends on it.
}
\label{tab:ablation_appendix}
% \vspace{-1em}
\end{table}

\section{Analysis of Using the VGGT Backbone}

\begin{table}[t]
\renewcommand{\thetable}{S\arabic{table}}
\centering
\resizebox{0.44\textwidth}{!}{
\begin{tabular}{lcccc}
\toprule
Method & PSNR$\uparrow$ & SSIM$\uparrow$ & LPIPS$\downarrow$ & Time(s)$\downarrow$ \\
\midrule
with VGGT~\cite{wang2025vggt} & \textbf{28.81} & \textbf{0.907} & \textbf{0.108} & 0.815 \\
Ours      & 28.56 & 0.901 & 0.110 & \textbf{0.337}\\
\bottomrule
\end{tabular}}
\caption{\textbf{Comparisons with and without the VGGT Backbone.} Using a pre-trained VGGT as the feature and depth extractor slightly improves performance but requires more inference time. Experiments are conducted on the RealEstate10K dataset
at a resolution of 360$\times$640 with 12 input views.}
\label{tab:vggt_appendix}
\end{table}

We also considered using VGGT~\cite{wang2025vggt} as the backbone for feature extraction instead of the depth-anything-v2-small~\cite{yang2024depth}. Since VGGT also utilizes DINOv2~\cite{oquabdinov2} and DPT head~\cite{ranftl2021vision} structures, we fused the global and geometric features extracted from VGGT. The experiment was also conducted on the RealEstate10K dataset with 12 input views. As shown in Table~\ref{tab:vggt_appendix}, using a pre-trained VGGT backbone slightly improves reconstruction quality (higher PSNR/SSIM and lower LPIPS), likely due to VGGT being trained on a larger dataset. However, the inference time is significantly longer when using VGGT, as it has many more model parameters (1B) compared to depth-anything-v2-small (24.8M). As a result, we continue to adopt the depth-anything-v2-small backbone. This experiment also demonstrates that our framework is not limited to a specific feature extractor.

\section{Supplementary Video}
We provide a supplementary video showcasing additional visual results rendered from multiple viewpoints. We highly recommend watching it to better appreciate the view consistency and high fidelity achieved by our method.

\section{Limitations and Future Work}

Although our proposed method demonstrates improvements in feed-forward 3D Gaussian Splatting for novel view synthesis, there are still certain limitations that could be addressed in future work. One key limitation of our model is that, although it is efficient, it may still face difficulties when processing very high-resolution datasets (e.g., those exceeding 1K), where fine details are not always accurately captured due to the inherent trade-off between model complexity and memory constraints. 
Moreover, our current Z-order transformer employs one or two Z-order blocks to aggregate Gaussian primitives. Using more such blocks can further reduce the number of Gaussian points; however, it leads to a noticeable degradation in performance, as illustrated in Fig.~7 of the main paper. Therefore, achieving higher compression without significant performance loss remains a challenging task.

In future work, several possible directions could be explored to further improve the proposed approach. One potential direction is to investigate hierarchical or multi-scale feature representations, which might help preserve fine-grained details when processing very high-resolution datasets while maintaining computational efficiency. 
It is also worth exploring alternative designs of the Z-order transformer, such as varying block configurations or introducing learnable aggregation depths to achieve a better balance between Gaussian primitive reduction and performance. Finally, combining the proposed framework with hybrid neural rendering methods~\cite{muller2022instant,fang2025nerf,wang2025hyrf,lin2025hybridgs} might help enhance robustness and generalization across diverse scenarios.

\end{document}